\documentclass{article} 
\usepackage{iclr2023_conference_tinypaper,times}

\usepackage{amsmath,amsfonts,bm}

\def\eqref#1{equation~\ref{#1}}

\def\1{\bm{1}}

\DeclareMathAlphabet{\mathsfit}{\encodingdefault}{\sfdefault}{m}{sl}
\SetMathAlphabet{\mathsfit}{bold}{\encodingdefault}{\sfdefault}{bx}{n}

\newcommand{\E}{\mathbb{E}}

\newcommand{\R}{\mathbb{R}}

\DeclareMathOperator*{\argmax}{arg\,max}

\usepackage{hyperref}
\usepackage{url}
\usepackage{algorithm}
\usepackage{graphicx}
\usepackage{wrapfig}
\usepackage{booktabs}

\usepackage{algpseudocode}

\usepackage{etoolbox}
\makeatletter
\def\thanks#1{\protected@xdef\@thanks{\@thanks
        \protect\footnotetext{#1}}}
\makeatother

\title{Toward Computationally Efficient Inverse Reinforcement Learning via Reward Shaping}

\iclrfinalcopy 

\author{Lauren H. Cooke*, Harvey Klyne*, Edwin Zhang* \thanks{*Equal contribution.  First author order is alphabetical.} \\
Harvard University\\
\And
Cassidy Laidlaw \\
University of California, Berkeley \\
\AND
Milind Tambe, Finale Doshi-Velez  \\
Harvard University
}

\begin{document}

\maketitle

\begin{abstract}
Inverse reinforcement learning (IRL) is computationally challenging, with common approaches requiring the solution of multiple reinforcement learning (RL) sub-problems. This work motivates the use of potential-based reward shaping to reduce the computational burden of each RL sub-problem. This work serves as a proof-of-concept and we hope will inspire future developments towards computationally efficient IRL.
\end{abstract}

\section{Introduction}

Inverse reinforcement learning (IRL) is the task of deriving a reward function that recovers expert behavior within an environment \citep{ngAlgorithms2000} and can be computationally expensive to solve. IRL algorithms typically consist of a loop in which every step requires finding the optimal policy for the current reward estimate (e.g.  \cite{abbeel_ng_2004, RamachandranBIRL, wulfmeier2016maximum}).
 This means that within a single IRL optimization multiple reinforcement learning (RL) problems need to be solved, each of which may be challenging. One can solve RL tasks by planning actions sufficiently far into the future \citep{sutton2018reinforcement}, and the necessary planning depth is a measure of the computational challenge of the problem. In the special case where the RL optimization makes use of a sample-based solver, planning depth can be thought of in terms of sample complexity \citep{kakade2003sample}. Previous works have attempted to reduce the overall cost of IRL by deliberately truncating the planning depth, accepting an approximation to the optimal policy at each iteration \citep{macglashan2015between, xu2022receding}.

Since multiple reward functions can encode an optimal policy \citep{RussellIRL, CaoID}, we have some choice about what reward function we optimize at each iteration. We envision using potential-based reward shaping \citep{NgShaping} to reduce the computational cost of each RL sub-problem without altering any of the optimal policies. This itself is too large a goal for the present work, so we focus our efforts on demonstrating a proof-of-concept in a simplified setting. In particular, we examine how sample trajectories from both optimal and random policies may be used to select a potential function for an initial feasible reward (one which encodes the optimal policy), which we call planning-aware reward shaping. Previous work on reward shaping includes \cite{hu2020learning,dong2020lyapunov,cheng2021heuristicguided,gupta2022unpacking,de2023guaranteeing}. To be clear, our present procedure does not directly address the problem of making IRL more computationally efficient, but we hope that the conclusions drawn may inspire future work.

\section{Planning-aware reward shaping}
\label{sec:shaping}

Suppose we have been given a Markov Decision Process without a reward $\mathcal{M} \setminus \mathcal{R}=(\mathcal{S}, \mathcal{A}, \gamma, P)$, and using optimal trajectories have learned a feasible reward $R_0$ using some IRL algorithm. We also assume access to a set of trajectories which have selected actions uniformly at random, and we use this additional exploration information to make a one-step adjustment to $R_0$. This adjustment takes the form of a potential function $\Phi:\mathcal{S}\to \R$, with our final reward function estimate taking the form 
\begin{equation}
R_{\Phi}(s,a) := R_0(s, a) + \gamma \E_{s' \sim p(\cdot \mid s,a)}[\Phi(s')] - \Phi(s). \label{eqn:shaping}
\end{equation} The optimal policies for $R_0$ and $R_\Phi$ are the same for any $\Phi$, and any equivalent reward may be written in the form $R_\Phi$ for some $\Phi$ \citep{NgShaping}. Crucially, the planning depths associated with rewards $R_\Phi$ may differ across choices of potential function $\Phi$. We stress that the goal of this work is to inspire future investigation into computationally efficient IRL, a problem we do not claim to have solved here. Our goal is to choose a shaped reward $R_\Phi$ which minimizes a certain bound on an algorithm-agnostic measure of planning depth \citep{LaidlawEffectiveHorizon}. This bound has been found to be strongly correlated with the sample complexities of modern deep learning RL procedures --- including DQN \citep{mnih2015DQN} and PPO \citep{schulman2017PPO} --- across a range of tasks. Denote by $Q_\Phi^\text{rand}(s,a)$ the $Q$-function associated with the uniform-at-random policy $\pi^\text{rand}(a \mid s) := 1/|A|$, and write $V_\Phi^*(s)$ for the value function associated with the optimal policy $\pi^*$. Recall that the optimal policy $\pi^*$ does not depend on $\Phi$. 
Following the derivation in \autoref{app:derivation}, we find that our objective is minimized (potentially non-uniquely) by 
\begin{equation}
    \Phi(s) = \big\{\max_{a\in A} Q_{0}^\text{rand}(s,a) + V_{0}^*(s)\big\} / 2. \label{eqn:potential}
\end{equation}
Both $V_{0}^*$ and $Q_{0}^\text{rand}$ are learnable from the expert and random exploration trajectories respectively. However, assuming access to the optimal value function $V_0^*$ trivializes the forward RL challenge (e.g. set $R(s) = V_0^*(s)$ and act greedily over next actions). We anticipate that IRL algorithms may be able to iteratively update the shaping potential $\Phi$ based on the current estimates of $Q_{0}^\text{rand}$ and $V_{0}^*$, which may improve the overall computational efficiency. We further remark that estimating $Q_0^\text{rand}$ is much easier than estimating $V_0^*$, so shaping based on (\ref{eqn:potential}) might have better finite-sample performance than potentials based on estimates of $V_0^*$ alone.

\section{Experiments}
\label{sec:numerical}

\begin{wrapfigure}{r}{0.45\textwidth}
\vspace{-20pt}
    \centering
    \includegraphics[width=0.45\textwidth]{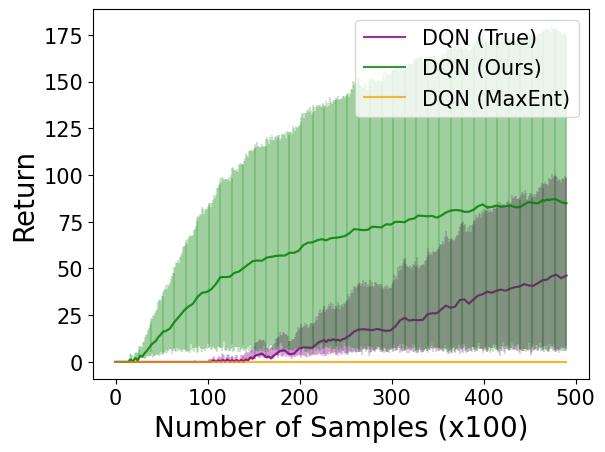}
    \vspace{-15pt}
    \caption{Return (measured by $R_0$) obtained by DQN. We compare learning with $R_0$ (purple), $R_\Phi$ (green), and $R_{\text{MaxEnt}}$ (orange), averaged across 500 random DQN seeds, each training for 50K steps. Our shaped reward $R_\Phi$ enables DQN to converge to the optimal value faster than the initial reward $R_0$, demonstrating a reduction of planning depth. DQN fails to optimize $R_{\text{MaxEnt}}$.}
    \label{fig:efficiency}
    \vspace{-11pt}
\end{wrapfigure}

We demonstrate that an oracle version of our procedure reduces the sample complexity for DQN (Figure~\ref{fig:efficiency}), which we use as a proxy for planning depth. We evaluate our method in a $5\times 5$ deterministic gridworld since we can easily find the optimal policy, yet DQN struggles and takes tens of thousands of steps to converge (\citet[Tab.~G.4]{LaidlawEffectiveHorizon}). 

In these experiments, we first fix transition dynamics and an initial reward $R_0$ before solving for the optimal policy $\pi^*$ by value iteration. As an IRL baseline, we also perform Maximum Entropy IRL \citep{ZiebartMaxEnt} on $\pi^*$ to obtain a reward $R_\text{MaxEnt}$. We compute the optimal value function $V_0^*$ and the random policy $Q$-function $Q^\text{rand}_0$ using Monte Carlo, the potential function $\Phi$ using (\ref{eqn:potential}), and the shaped reward $R_\Phi$ using (\ref{eqn:shaping}). Note that $R_0$, $R_\Phi$, and $R_\text{MaxEnt}$ all encode the same optimal policy. We compare the planning depths of these three rewards using the sample efficiency of DQN, finding that the shaped reward $R_\Phi$ enables DQN to converge to the optimal solution faster than the initial reward $R_0$. We also find that DQN fails to optimize $R_{\text{MaxEnt}}$, with the policy getting stuck in a local optimum state rather than reaching the goal state. 
Implementation details and code to reproduce our experiments are included in \autoref{app:implementation} and the supplementary materials.

\section{Conclusion}

In this work, we motivate planning-aware reward shaping to reduce the computational complexity of IRL. We demonstrate that an oracle version of our procedure reduces the planning depth of an RL task, as measured by the sample complexity of DQN (Section~\ref{sec:numerical}). Compared to existing IRL approaches, we leverage the additional information included in random trajectories to apply automatic reward shaping. Our hope is that our procedure may inspire novel IRL algorithms which are more computationally efficient. While we focus on the IRL setting, adaptive shaping procedures such as ours may also be of interest to the broader RL community.


\bibliography{iclr2023_conference_tinypaper}

\begin{thebibliography}{21}
\providecommand{\natexlab}[1]{#1}
\providecommand{\url}[1]{\texttt{#1}}
\expandafter\ifx\csname urlstyle\endcsname\relax
  \providecommand{\doi}[1]{doi: #1}\else
  \providecommand{\doi}{doi: \begingroup \urlstyle{rm}\Url}\fi

\bibitem[Abbeel \& Ng(2004)Abbeel and Ng]{abbeel_ng_2004}
Pieter Abbeel and Andrew~Y. Ng.
\newblock Apprenticeship learning via inverse reinforcement learning.
\newblock In \emph{Proceedings of the Twenty-First International Conference on Machine Learning}, pp.\  1--8, 2004.

\bibitem[Cao et~al.(2021)Cao, Cohen, and Szpruch]{CaoID}
Haoyang Cao, Samuel~N. Cohen, and Łukasz Szpruch.
\newblock Identifiability in inverse reinforcement learning.
\newblock In \emph{Proceedings of the 35th Conference on Neural Information Processing Systems}, pp.\  1--11, 2021.

\bibitem[Cheng et~al.(2021)Cheng, Kolobov, and Swaminathan]{cheng2021heuristicguided}
Ching-An Cheng, Andrey Kolobov, and Adith Swaminathan.
\newblock Heuristic-guided reinforcement learning.
\newblock \emph{arXiv}, pp.\  2106.02757, 2021.

\bibitem[{De Lellis} et~al.(2023){De Lellis}, Coraggio, Russo, Musolesi, and di~Bernardo]{de2023guaranteeing}
Francesco {De Lellis}, Marco Coraggio, Giovanni Russo, Mirco Musolesi, and Mario di~Bernardo.
\newblock Guaranteeing control requirements via reward shaping in reinforcement learning.
\newblock \emph{arXiv}, pp.\  2311.10026, 2023.

\bibitem[Dong et~al.(2020)Dong, Tang, and Yuan]{dong2020lyapunov}
Yunlong Dong, Xiuchuan Tang, and Ye~Yuan.
\newblock {Principled reward shaping for reinforcement learning via Lyapunov stability theory}.
\newblock \emph{Neurocomputing}, 393:\penalty0 83--90, 2020.

\bibitem[Gupta et~al.(2022)Gupta, Pacchiano, Zhai, Kakade, and Levine]{gupta2022unpacking}
Abhishek Gupta, Aldo Pacchiano, Yuexiang Zhai, Sham~M. Kakade, and Sergey Levine.
\newblock Unpacking reward shaping: Understanding the benefits of reward engineering on sample complexity.
\newblock \emph{arXiv}, pp.\  2210.09579, 2022.

\bibitem[Hu et~al.(2020)Hu, Wang, Jia, Wang, Chen, Hao, Wu, and Fan]{hu2020learning}
Yujing Hu, Weixun Wang, Hangtian Jia, Yixiang Wang, Yingfeng Chen, Jianye Hao, Feng Wu, and Changjie Fan.
\newblock {Learning to Utilize Shaping Rewards: A New Approach of Reward Shaping}.
\newblock \emph{arXiv}, pp.\  2011.02669, 2020.

\bibitem[Kakade(2003)]{kakade2003sample}
{Sham M.} Kakade.
\newblock \emph{On the Sample Complexity of Reinforcement Learning}.
\newblock PhD thesis, University College London, 2003.

\bibitem[Kingma \& Ba(2014)Kingma and Ba]{kingma2014adam}
Diederik~P Kingma and Jimmy Ba.
\newblock Adam: A method for stochastic optimization.
\newblock \emph{arXiv}, pp.\  1412.6980, 2014.

\bibitem[Laidlaw et~al.(2023)Laidlaw, Russell, and Dragan]{LaidlawEffectiveHorizon}
Cassidy Laidlaw, Stuart Russell, and Anca Dragan.
\newblock {Bridging RL Theory and Practice with the Effective Horizon}.
\newblock \emph{arXiv}, pp.\  2304.09853, 2023.

\bibitem[MacGlashan \& Littman(2015)MacGlashan and Littman]{macglashan2015between}
James MacGlashan and {Michael L.} Littman.
\newblock Between imitation and intention learning.
\newblock In Qiang Yang and Michael Wooldridge (eds.), \emph{Proceedings of the Twenty-Fourth International Joint Conference on Artificial Intelligence}, pp.\  3692--3698, 2015.

\bibitem[Mnih et~al.(2015)Mnih, Kavukcuoglu, Silver, Rusu, Veness, Bellemare, Graves, Riedmiller, Fidjeland, Ostrovski, Petersen, Beattie, Sadik, Antonoglou, King, Kumaran, Wierstra, Legg, and Hassabis]{mnih2015DQN}
Volodymyr Mnih, Koray Kavukcuoglu, David Silver, {Andrei A.} Rusu, Joel Veness, {Marc G.} Bellemare, Alex Graves, Martin Riedmiller, {Andreas K.} Fidjeland, Georg Ostrovski, Stig Petersen, Charles Beattie, Amir Sadik, Ioannis Antonoglou, Helen King, Dharshan Kumaran, Daan Wierstra, Shane Legg, and Demis Hassabis.
\newblock Human-level control through deep reinforcement learning.
\newblock \emph{Nature}, 518:\penalty0 529--533, 2015.

\bibitem[Ng \& Russell(2000)Ng and Russell]{ngAlgorithms2000}
{Andrew Y.} Ng and {Stuart J.} Russell.
\newblock Algorithms for inverse reinforcement learning.
\newblock In \emph{Proceedings of the Seventeenth International Conference on Machine Learning}, pp.\  663--670, 2000.

\bibitem[Ng et~al.(1999)Ng, Harada, and Russell]{NgShaping}
{Andrew Y.} Ng, {Daishi} Harada, and {Stuart J.} Russell.
\newblock Policy invariance under reward transformations: Theory and application to reward shaping.
\newblock In \emph{Proceedings of the Sixteenth International Conference on Machine Learning}, pp.\  278--287, 1999.

\bibitem[Ramachandran \& Amir(2007)Ramachandran and Amir]{RamachandranBIRL}
Deepak Ramachandran and Eyal Amir.
\newblock Bayesian inverse reinforcment learning.
\newblock In \emph{Proceedings of the 20th International Joint Conference on Artificial Intelligence}, pp.\  2586--2591, 2007.

\bibitem[Russell(1998)]{RussellIRL}
Stuart Russell.
\newblock {Learning Agents for Uncertain Environments (Extended Abstract)}.
\newblock In \emph{Proceedings of the Eleventh Annual Conference on Computational Learning Theory}, pp.\  101--103. Association for Computing Machinery, 1998.
\newblock ISBN 1581130570.

\bibitem[{Schulman} et~al.(2017){Schulman}, {Wolski}, {Dhariwal}, {Radford}, and {Klimov}]{schulman2017PPO}
John {Schulman}, Filip {Wolski}, Prafulla {Dhariwal}, Alec {Radford}, and Oleg {Klimov}.
\newblock Proximal policy optimization algorithms.
\newblock \emph{arXiv}, pp.\  1707.06347, 2017.

\bibitem[Sutton \& Barto(2018)Sutton and Barto]{sutton2018reinforcement}
R.S. Sutton and A.G. Barto.
\newblock \emph{Reinforcement Learning, second edition: An Introduction}.
\newblock Adaptive Computation and Machine Learning series. MIT Press, 2018.
\newblock ISBN 9780262352703.

\bibitem[Wulfmeier et~al.(2016)Wulfmeier, Ondruska, and Posner]{wulfmeier2016maximum}
Markus Wulfmeier, Peter Ondruska, and Ingmar Posner.
\newblock Maximum entropy deep inverse reinforcement learning.
\newblock \emph{arXiv}, pp.\  1507.04888, 2016.

\bibitem[Xu et~al.(2022)Xu, Gao, and Hsu]{xu2022receding}
Yiqing Xu, Wei Gao, and David Hsu.
\newblock Receding horizon inverse reinforcement learning.
\newblock In {Alice H.} Oh, Alekh Agarwal, Danielle Belgrave, and Kyunghyun Cho (eds.), \emph{Proceedings of the Thirty-Sixth Conference on Neural Information Processing Systems}, 2022.

\bibitem[Ziebart et~al.(2008)Ziebart, Maas, Bagnell, and Dey]{ZiebartMaxEnt}
Brian~D. Ziebart, Andrew Maas, J.~Andrew Bagnell, and Anind~K. Dey.
\newblock Maximum entropy inverse reinforcement learning.
\newblock In Anthony Cohn (ed.), \emph{Proceedings of the 23rd National Conference on Artificial Intelligence}, pp.\  1433--1438, 2008.

\end{thebibliography}
\bibliographystyle{iclr2023_conference_tinypaper}

\appendix

\section*{Appendix}

\section{Reward shaping optimization objective}
\label{app:derivation}

\cite{LaidlawEffectiveHorizon} consider algorithm-agnostic proxies for the sample complexities of modern deep RL approaches across measures of correlation, tightness, and accuracy. 
They introduce the effective horizon $H := \min_{k} k + \log_{|A|} m_k$, where $k$ is a tuning parameter in a simple Monte Carlo algorithm (see \citet[Alg.~1]{LaidlawEffectiveHorizon}) and $m_k$ is the minimum sample size this algorithm requires to find an optimal policy with probability at least $1/2$. It is not feasible to compute $H$ in closed form, but they find that a particular bound serves as a good proxy (\citet[Thm.~5.4]{LaidlawEffectiveHorizon}). Considering an MDP with finite-time horizon $T$ and $\gamma=1$, it holds that:
\begin{equation*}
    H \leq \min_{k = 1,\ldots, T} k + \max_{t \in T, s \in S, a \in A} \log_{|A|}\left(\frac{Q_t^k(s,a)V_t^*(s)}{\Delta_t^k(s)^2}\right) + \log_A6\log(2T|A|^k),
\end{equation*}
where \begin{equation*}
    \Delta_t^k(s) = \max_{a\in A} Q^k_t(s,a) - \max_{a'\notin \argmax_{a} Q^k_t(s,a)} Q^k_t(s,a'),
\end{equation*}
and $Q^k_t$, $V^*_t$, and $\Delta^k_t$ are defined as in \cite{LaidlawEffectiveHorizon}. 

We relax this bound by fixing $k=1$ --- which we think is reasonable considering how $k$ is fixed to $1$ in practice (e.g. \citet[Sec.~F.1]{LaidlawEffectiveHorizon}) --- but the potential function $\Phi$ we derive generalizes to other choices of $k$. We consider MDPs with $T=\infty, \gamma < 1$ and time-invariant policies, motivating the following optimization for our planning-aware reward shaping:
\begin{equation}\label{eqn:obj}
    \max_{\Phi:\mathcal{S}\to \R} \bigg\{ \max_{s\in\mathcal{S}, a\in A} \log_{|A|} \bigg( \frac{Q_{\Phi}^\text{rand}(s,a) V_{\Phi}^*(s)}{\Delta_{\Phi}(s)^2} \bigg)\bigg\}.    
\end{equation}

This is strictly increasing in the following criteria:
\begin{equation*}
\ell(\Phi; R_0) := 
\max_{s \in \mathcal{S}, a \in A} \frac{Q_{\Phi}^\text{rand}(s,a) V_{\Phi}^*(s)}{\Delta_{\Phi}(s)^2}.
\end{equation*}

 In fact, for any policy $\pi$ the associated $Q$-function and value function transform linearly under reward shaping \citep[Cor.~2]{NgShaping}:\begin{equation*}
     Q^{\pi}_{\Phi}(s,a) = Q^{\pi}_0(s,a)-\Phi(s);\quad V^{\pi}_{\Phi}(s) = V^\pi_0(s)-\Phi(s).
 \end{equation*}
 Therefore $\Delta_\Phi(s)=\Delta_0(s)$ for all potentials $\Phi$, so the objective reduces to
\begin{equation*}
     \ell(\Phi; R_0) = \max_{s \in \mathcal{S}, a \in A} \frac{\big\{Q_{0}^\text{rand}(s,a) - \Phi(s)\big\}\big\{V_{0}^*(s) - \Phi(s)\big\}}{\Delta_{0}(s)^2}.
 \end{equation*}
The potential function (\ref{eqn:potential}) solves this quadratic for every $s\in\mathcal{S}$, and is thus a global minimizer. The solution can be found by straightforwardly taking the derivative of \autoref{eqn:obj} and setting to $0$.

\section{Implementation details for experiments}
\label{app:implementation}

In \autoref{fig:efficiency} we plot the returns achieved by each optimization procedure at each training episode by taking an average across 500 random seeds, along with 95\% bootstrapped confidence intervals. Each seed determined one individual training process, wherein we train for 500 episodes of 100 steps each, for a total of 50K training steps. Returns are evaluated at the end of each episode with respect to the initial reward function $R_0$, regardless of which reward function is used during training. This ensures that returns are comparable between objectives. We set a $100$ timestep limit on the environment.

\begin{table}[h]
    \centering
    \begin{tabular}{lll}
        \toprule 
        & Hyperparameter & Value \\
        \midrule
        DQN HP & Optimizer & Adam~\citep{kingma2014adam} \\
        & Critic architecture & MLP \\
        & Critic learning rate & 1{e}-3 \\
        & Critic hidden layers & 1 \\
        & Critic hidden dim & 24 \\
        & Critic activation function & 
        ReLU \\

        & Mini-batch size & 1024 \\
        & Number of gradient steps & 50K \\
        & Discount factor & 0.99 \\
        & Target update rate & 1 \\
        & Target update period & 8 \\
        & Loss Function & Huber Bellman Error \\
    \end{tabular}
    \caption{Hyperparameters for the DQN algorithm used in Section~\ref{sec:numerical}.}
    \label{tab:dqn-hp}.
\end{table}

To perform DQN we use an MIT-licensed implementation (\url{github.com/mswang12/minDQN}) with hyperparameters as in \autoref{tab:dqn-hp}. Code to reproduce our experiments is included in the supplementary materials.

\end{document}